\documentclass{article}
\usepackage{arxiv}

\usepackage{times}

\usepackage{soul}
\usepackage{url}
\usepackage{authblk}
\usepackage[T1]{fontenc}    
\usepackage{hyperref}
\usepackage[utf8]{inputenc}
\usepackage[small]{caption}
\usepackage{graphicx}
\usepackage{amsmath}
\usepackage{booktabs}
\usepackage{subcaption}

\usepackage{array}
\newcolumntype{L}[1]{>{\raggedright\let\newline\\\arraybackslash\hspace{0pt}}p{#1}}
\newcolumntype{C}[1]{>{\centering\let\newline\\\arraybackslash\hspace{0pt}}p{#1}}
\newcolumntype{R}[1]{>{\raggedleft\let\newline\\\arraybackslash\hspace{0pt}}p{#1}}

\usepackage{amsfonts,amssymb}       
\usepackage{nicefrac}       
\usepackage{microtype}
\usepackage{algorithmic}
\usepackage[ruled,vlined]{algorithm2e}
\usepackage{caption} 

\title{Dynamic Partial Removal: A Neural Network Heuristic for Large Neighborhood Search}

\author{
  Mingxiang Chen \qquad Lei Gao \qquad Qichang Chen \qquad  Zhixin Liu  \thanks{liuzhixin@watermirror.ai}
}
\affil{WaterMirror Inc. \\Shenzhen, China}

\begin{document}

\maketitle

\begin{abstract}

This paper presents a novel neural network design that learns the heuristic for Large Neighborhood Search (LNS). LNS consists of a destroy operator and a repair operator that specify a way to carry out the neighborhood search to solve the Combinatorial Optimization problems. 
The proposed approach in this paper applies a Hierarchical Recurrent Graph Convolutional Network (HRGCN) as a LNS heuristic, namely Dynamic Partial Removal, with the advantage of adaptive destruction and the potential to search across a large scale, as well as the context-awareness in both spatial and temporal perspective. This model is generalized as an efficient heuristic approach to different combinatorial  optimization problems, especially to the problems with relatively tight constraints. We apply this model to vehicle routing problem (VRP)  in this paper as an example. The experimental results show that this approach outperforms the traditional LNS heuristics on the same problem as well. The source code is available at \href{https://github.com/water-mirror/DPR}{https://github.com/water-mirror/DPR}.

\end{abstract}

\section{Introduction}

The combinatorial optimization (CO) problems and their solution techniques are widely applied in a number of real-world industries such as manufacturing, supply chain, retailing, etc. Combinatorial optimization is targeting at finding the value of the decision variable $x^\ast$ from a finite set $\mathcal{X}$ that is \textit{feasible}  and that optimizes the objective function, where the objective function is usually a cost function or a loss function to be minimized, or a value function to be maximized. Here a \textit{feasible} solution $x\in\mathcal{X}$ is the value of decision variable that satisfies the problem's constraints. In practical problems, the decision variable $x$ is generally with high dimensionality, making the scale of $\mathcal{X}$ exponentially large.  Consequentially the  algorithms on solving combinatorial optimization problems, especially the ones with large scale, is of  great interest of research for decades. To solve the large-scale problems within a limited computing time period, heuristic and meta-heuristic methods are developed to achieve a good  balance between computational time and solution quality, among which the Large Neighborhood Search (LNS) is a popular one due to its close-to-optimal solution quality and fast converging speed. LNS can be applied together with other meta-heuristic methods such as Tabu search and Simulated Annealing for better performance.

\begin{figure}
    \centering
    \includegraphics[width=.95\linewidth]{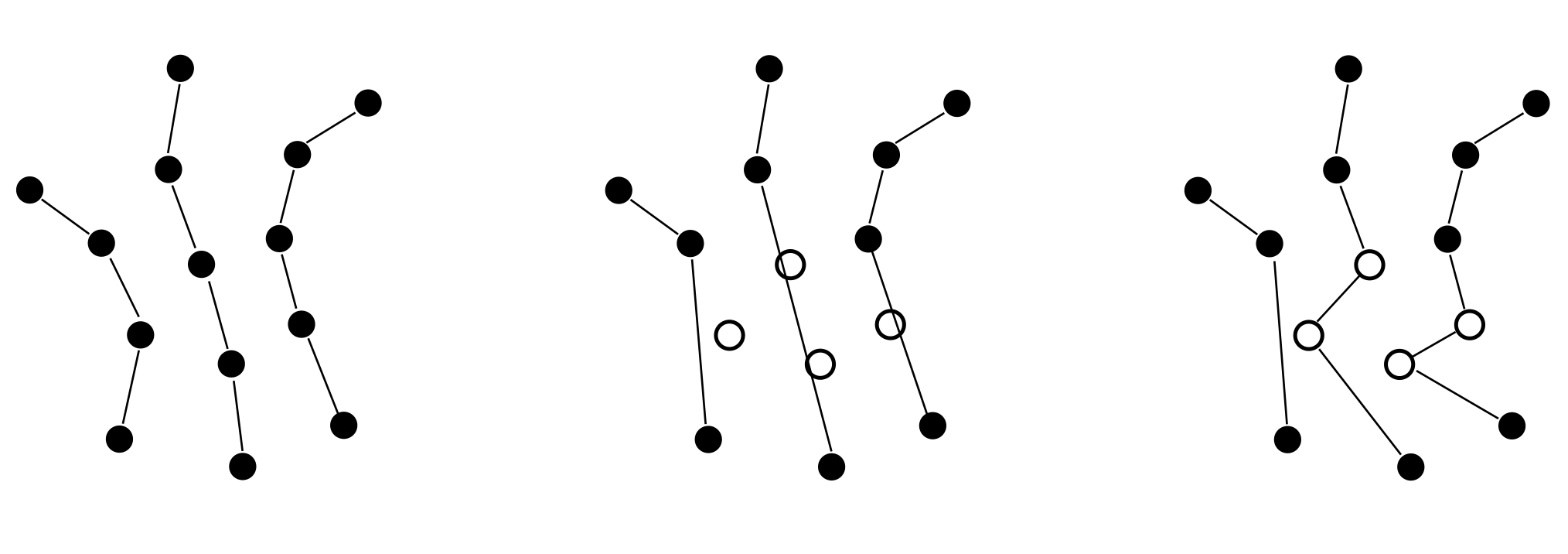}
    \caption{An example of large neighborhood search. From left to right: the original solution; a destroyed one; a repaired one.}\label{Fig:lns}
\end{figure}

LNS roots from Neighborhood Search heuristics. Neighborhood search algorithms try to improve the initial feasible solution  by repeatedly searching within a \textit{neighborhood} of current solution, where \textit{neighborhood} is defined as a set of feasible solutions that are similar to the current one. The neighborhood of solution $x$ is described mathematically as: $N(x):\{x^\prime=\mathcal{N}(x)\}$ where $\mathcal{N}$ is the exploration function to traverse the similar feasible solutions $x^\prime$ from current solution $x$.  The choice of neighborhood from where the new solution is chosen is apparently essential for better solution quality and for better exploration effectiveness. LNS was introduced in \cite{shaw1998using} to extend the  neighborhood, because a larger neighborhood yields a better result by preventing algorithms from getting trapped in local optima. However it is time-consuming to search a larger neighborhood thoroughly. This fact poses a challenge on the design of LNS heuristics which is capable to converge to a solution of good quality within a reasonable amount of computing time. LNS tackles this challenge by expanding the neighborhood into $N(x):\{x^\prime=\mathcal{N}(x^{\prime\prime}), x^{\prime\prime}\in x\bigcup N(x)\}$, meaning that an updated solution $x^\prime$ is accepted (i.e., updating to $x^\prime$ from $x^{\prime\prime}$) at each iteration even if it does not improve the solution quality. In this way the search in a relatively larger neighborhood is implemented in an iterative manner. Meanwhile the exploration function $\mathcal{N}$ has been replaced into the combination of two consecutive operators: a destroy operator and a repair operator, following the idea that part of the current solution is destroyed and then is repaired afterwards. Using Vehicle Routing Problem (VRP) as an example, the destroy operator eliminates a portion of customer nodes from current routes, then the repair operator inserts the removed nodes back into the alternative or the same routes. This process is shown as in Fig \ref{Fig:lns}.

The key aspect of LNS is the design of operators, with consideration of both the computational time and the search efficiency. The most commonly used destroy operators include random customer removal, random route removal, related route removal or workday level operators. Typical repair operators include least cost heuristic and regret based heuristic. Besides the design of operators, the degree of destruction is one of the most important parameters to be determined. There are different strategies of setting the degree of destruction, for example, a constant degree, or a random degree, or an adaptive degree depending on the current solution. In recent years, multi-level approaches such as Adaptive Large Neighborhood Search (ALNS) \cite{ropke2006adaptive} and Slack Induction by String Removals (SISR) \cite{christiaens2020slack} are introduced to improve the searching efficiency and effectiveness. The main takeaway of ALNS is that the combination of multiple heuristics is generally better than a single heuristic, and the weights of the multiple heuristics shall be adaptive. The main takeaway of SISR is that clustering of removed nodes rather than scattering is always preferable. These design considerations are the major inspirations of our  research work proposed in this paper. 

Another family of approaches to solve the combinatorial optimization problems is based on the deep learning and the reinforcement learning models. Researchers have been proposing various deep learning algorithms recently. Among them, the pointer network \cite{vinyals2015pointer}, which is inspired by the Seq2seq \cite{sutskever2014sequence} model and the attention mechanism\cite{graves2014neural}\cite{bahdanau2014neural}, is one of the earliest attempts. Later, more approaches have been proposed, which can be grouped into two categories in general. One category is to train a neural network model end-to-end to produce  a solution directly, by adopting the Encoder-Decoder framework or attention structures \cite{nazari2018reinforcement}\cite{kool2018attention}. Another paradigm focuses on solving the problems with iterative improvements like many heuristic search algorithms \cite{chen2019learning}\cite{gao2020learn}. 

We focus on the second paradigm, say, by applying a neural network heuristic to improve the solution repeatedly, because of its guarantee to the solution quality and the explainability of the network model. Several insights of the optimization problems and their heuristic techniques are to be examined for a reasonable design of the network. The first insight is that an excessive amount of combinatorial optimization problems can be expressed by a Graph $\mathcal{G}=(\mathcal{N}, \mathcal{A})$, where $\mathcal{N}$ represents the nodes  and $\mathcal{A}$ represents the arcs. For example, Network Flow problem, Vehicle Routing problems, Generalized Assignment problems, etc, can be depicted in this way. The recent developing on Graph Neural Network \cite{scarselli2008graph} can be leveraged to the network design for this purpose due to its power on  the information embedding and belief propagation of the graph topology \cite{khalil2017learning}. Other insights on the heuristic techniques include the considerations on the operators' design such as the degree of destruction, the adaptive operators, and the clustering of removed nodes in the graph.

Following the insights and considerations, we propose a novel network design for LNS heuristics on combinatorial optimization problems that can be depicted by a graph $\mathcal{G}$. The proposed adaptive destroy operator for LNS is called Dynamic Partial Removal (DPR) by introducing and applying the Hierarchical Recurrent Graph Convolutional Network (HRGCN). We use capacitatied VRP with time windows (CVRPTW) as an example to show how this approach works and its experimental results comparing to existing LNS algorithms. This approach can be easily applied to other variants of VRP, and to other combinatorial problems as well. 
Our approach differs from other deep-learning-based algorithms in the following aspects: a) this model can solve VRP with tight constraints, e.g. CVRPTW, which is described as ``a challenging task" in some recent papers \cite{nazari2018reinforcement}; b) this approach is able to be generalized to mixed-scale problems; c) this approach has the bandwidth to solve large-scaled VRP and its variants efficiently with as many as 800 customer nodes or even more.

This paper is organized as below. Our method along with the problem definition, the new design of heuristic search (DPR) and the structure of the neural network (HRGCN) are explained in Section 2. The experiment settings, hyper-parameters, and results are presented in Section 3. The final conclusion and future works are discussed in Section 4.

\section{Proposed Model}

\subsection{Problem Definition}

VRP is an integer programming problem that defines a group of fleet routes to serve a set of customer nodes. The solution is represented by a directed graph $\mathcal{G}=(\mathcal{N},\mathcal{A})$, where the nodes of the graph $i\in\mathcal{N}=\{0,1,...,N\}$ represents the customers to be served if $i>0$ or the depot if $i=0$. Each customer $i>0$ is associated with a demand (a number of goods to be delivered) $q_i$. Each arc $a_{i,j}\in\mathcal{A}$ represents a segment of a route that connects node $i$ and $j$. In capacitated vehicle routing problems with time windows (CVRPTW), vehicles have limited carrying capacities, and each customer has a service time window $[s_i, e_i]$, as well as a service (or, handling) time period $t_i$. The duration for each route is limited by the depot's time window $[s_0, e_0]$, where $s_0$ is usually 0.

\subsection{Dynamic Partial Removal as a Destroy Heuristic}

\begin{figure}[ht]
    \centering
    \begin{subfigure}{0.2\linewidth}
        \centering
        \includegraphics[width=\linewidth]{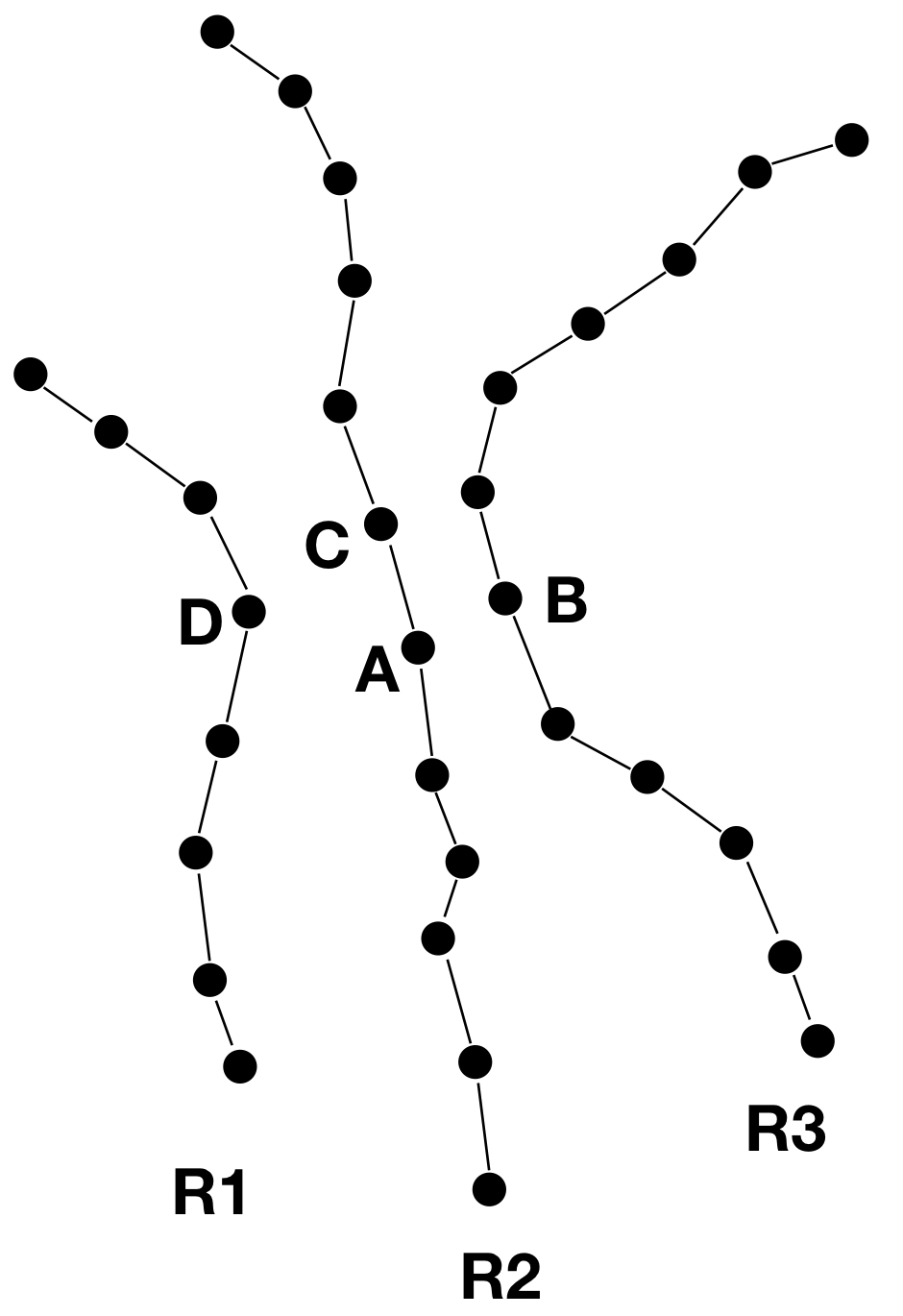}
        \caption{ } \label{fig:coeff_step_a}
    \end{subfigure}
    \hspace*{\fill} 
    \begin{subfigure}{0.2\linewidth}
        \centering
        \includegraphics[width=\linewidth]{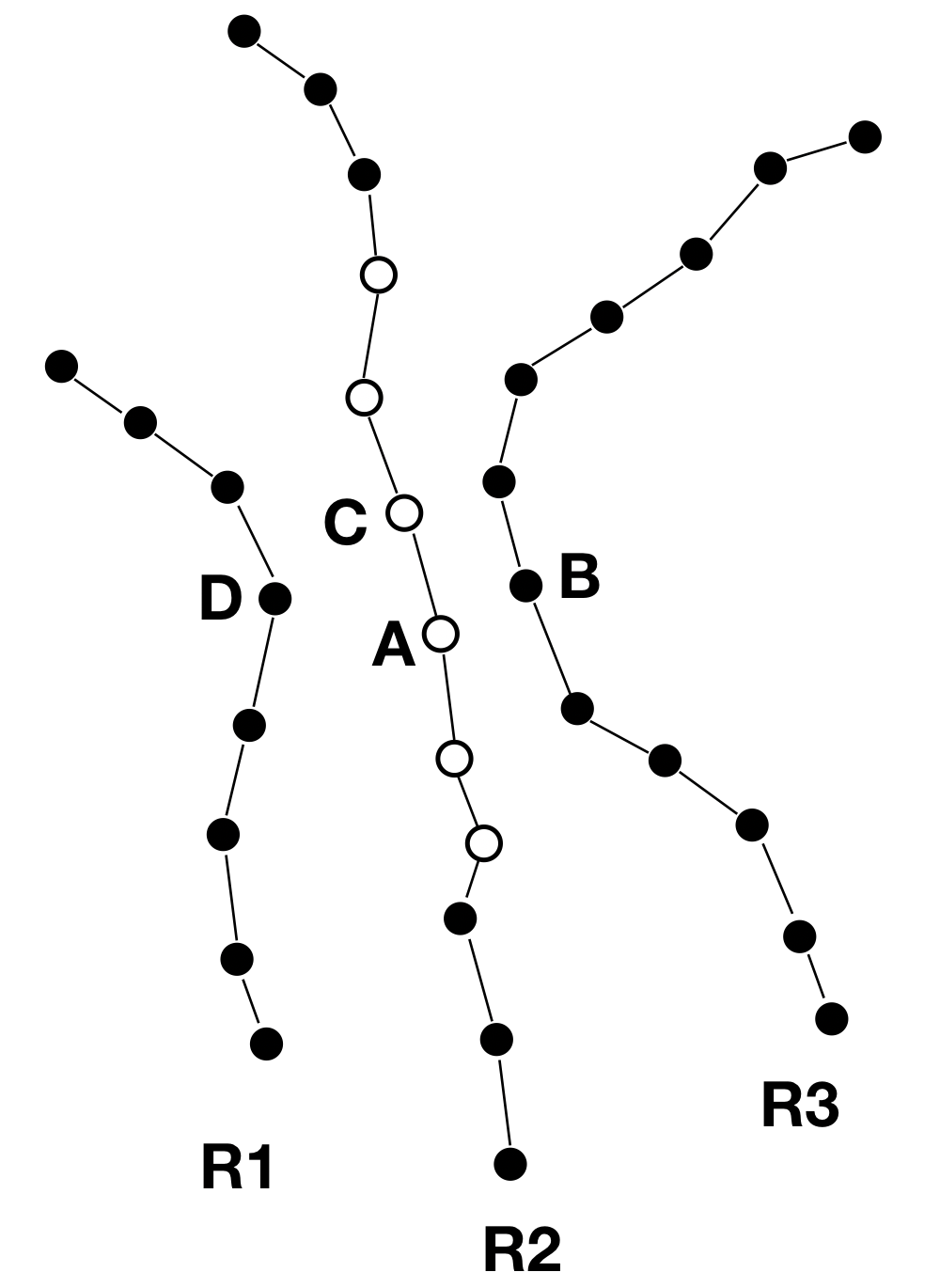}
        \caption{ } \label{fig:coeff_step_b}
    \end{subfigure}
    \hspace*{\fill} 
    \begin{subfigure}{0.2\linewidth}
        \centering
        \includegraphics[width=\linewidth]{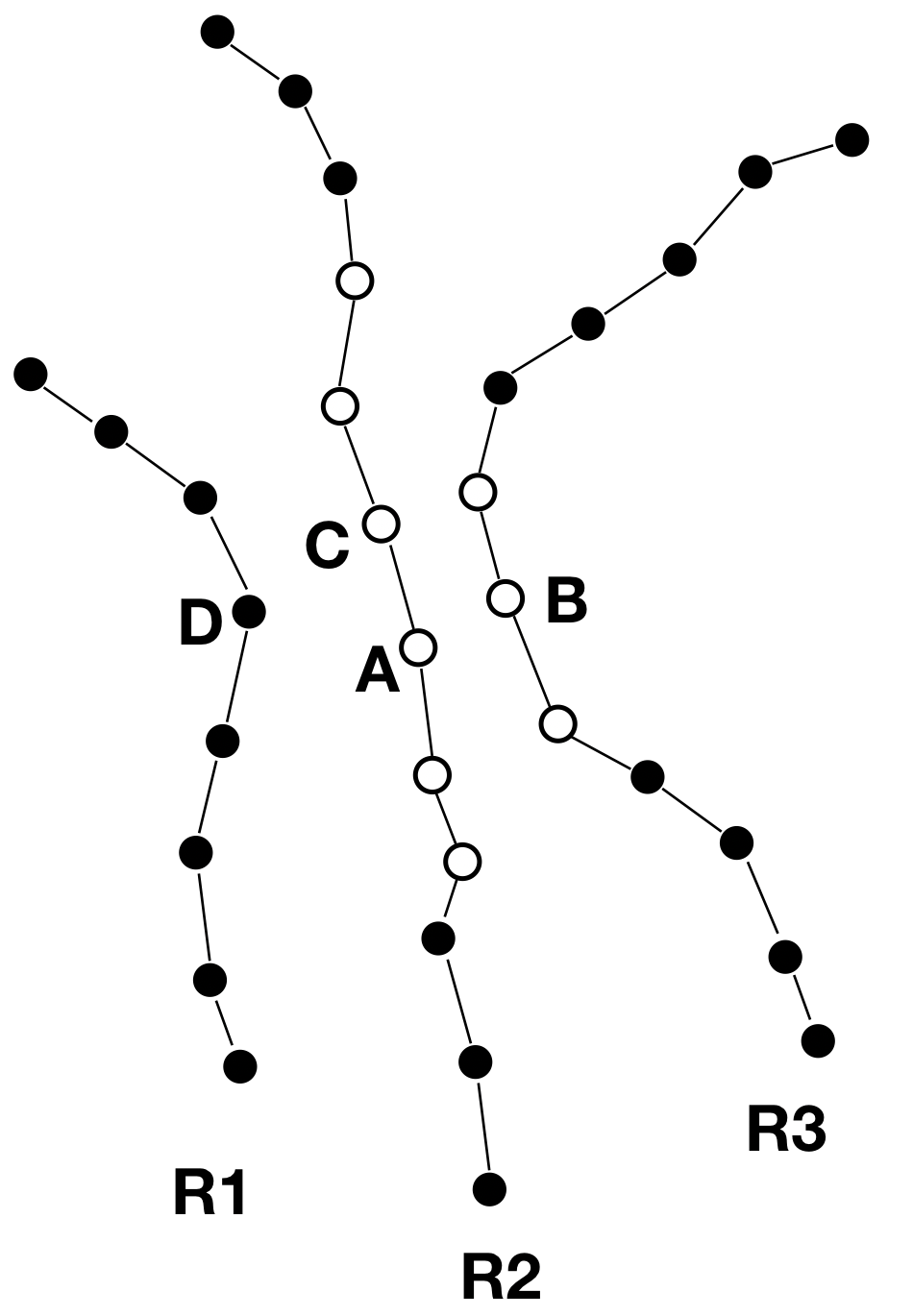}
        \caption{ } \label{fig:coeff_step_c}
    \end{subfigure}
    \hspace*{\fill} 
    \begin{subfigure}{0.2\linewidth}
        \centering
        \includegraphics[width=\linewidth]{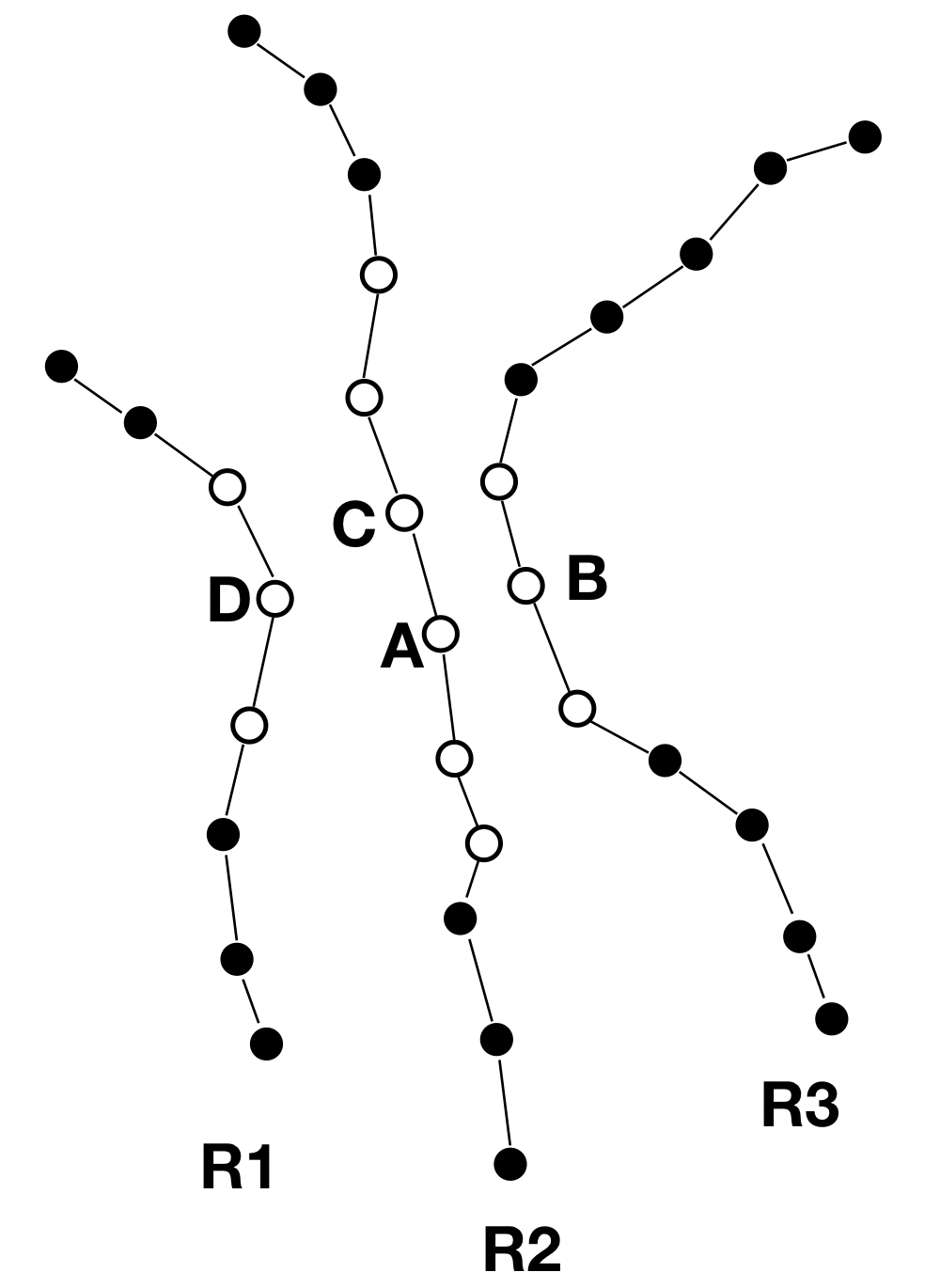}
        \caption{ } \label{fig:coeff_step_d}
    \end{subfigure}
    \caption{An illustration of DPR. Figure (a) is the original solution which includes three routes: R1, R2, and R3. The routes have 8 nodes, 12 nodes, and 12 nodes, respectively. The original anchor is node A, and the degree of destruction is 12. The four nearest neighbors of node A (including itself) are A, B, C, and D. Assume the route coefficients for these four nodes are 0.5, 0.25, 0.25, and 0.5, respectively. The destruction starts with the anchor node A: A belongs to R2 which has 12 nodes, and the coefficient is 0.5, therefore 6 out of 12 nodes are removed as shown in figure (b). The same logic is applied to node B since the degree of destruction (12) is larger than the accumulated number of removed nodes (figure (c)). Here node C is skipped since it is already purged. When the destruction comes to node D, the coefficient is 0.5 which means 4 out of 8 nodes should be removed along route R1, however the accumulated number of removed nodes is 9 up to this point, consequently only 3 nodes are to be removed from R1 and the DPR stops as degree of destruction is satisfied.} \label{fig:coeff_step}
\end{figure}

The destroy operator proposed in our algorithm is called "dynamic partial removal" (DPR). There are three parameters for this operator: an anchor node, the route coefficient, and the number of nodes to be removed.
Given the current solution at step $t$ as $x^{(t)}=[x^{(t)}_{r, n_r}]$, which is a matrix where the row $r$ indicates the $r$-th route, and the element at row $r$ and column $n_r$ indicates the $n$-th node on $r$-th route. The destroy operator is to remove a subset of elements $\{x^{(t)}_{r^\prime, n_r^\prime}\}$. This subset is clustered around the anchor node $\{x^{(t)}_{r^0, n_r^0}\}$. The route coefficient defines the allocation of this subset across different rows (i.e., routes in VRP), and the size of this subset determines the number of nodes to be removed, or the degree of destruction equivalently.  Figure \ref{fig:coeff_step} shows a concrete example of DPR, where the successive elements (nodes) on the route $r^0$ are removed with a maximal number determined by the route coefficient, and the same removal logic is applied to next closest row (route) if the degree of destruction is not satisfied, and so on so forth. Note that the depot (the $0^{th}$ node) will never be removed.  The similar approach can be applied to other combinatorial optimization problems where the decision variable $x$ is expressed in the form of a matrix. Please refer to figure \ref{fig:coeff_step} for details.

When the route coefficient is close to 1, the operator prefers a string removal pattern as to delete multiple consecutive nodes on the same path. If the route coefficient is close to 0, the operator tends to delete nodes with closest Euclidean distance.

We use the least-cost heuristic as the repair operator to repair the connections, by picking up the removed nodes at random order and inserting them at the feasible locations with minimum detour \cite{azi2014adaptive}. The algorithm (\ref{alg:DPRR}) shows the process of DPR, where $f(\theta_a)$ is the policy to pick up the anchor and $f(\theta_c)$ is the policy to output the route coefficients.

\begin{algorithm}[!htb]
    \caption{Dynamic Partial Removal}
    \label{alg:DPRR}
    \begin{algorithmic}
        \STATE \textbf{Input:} $x^{(t)}$
        \STATE Set $L_{destroy}$ as an empty list
        \STATE anchor $\leftarrow$ select an anchor according to $f(\theta_a)$
        \FOR{node $i$ in Neighbor(anchor)}
        \STATE Find the route $r$ to which the node $i$ belongs
        \STATE $removals \leftarrow removeSelected(r, i, f(\theta_c), L_{destroy})$
        \STATE $L_{destroy} \leftarrow L_{destroy} \cup removals$
        \IF{$len(L_{destroy}) = n_{destroy}$}
        \STATE \textbf{Break}
        \ENDIF
        \ENDFOR
        \STATE Remove the nodes in $L_{destroy}$ from $x^{(t)}$
        \STATE Shuffle the $L_{destroy}$
        \STATE Reinsert the nodes in $L_{destroy}$ into $x^{(t)}$
        \STATE \textbf{Output:} $x^{(t)}$
    \end{algorithmic}
\end{algorithm}

\subsection{HRGCN}

The details of the neural network that generates the above-mentioned anchor node and the route coefficients as well are explained in the following subsections in terms of node embedding, hierarchical network architecture, and the training method.

\subsubsection{Embeddings}

We embed each node into an 10-dimensional vector $n_i$, with each dimension as: 1) the node's $x$ coordinate, 2) the node's $y$ coordinate, 3) the customer's demand $q_i$, 4) the starting time of the service time window $s_i$, 5) the ending time of the service time window $e_i$, 6) the service time period $t_i$, 7) sum up of demands $q_i$ of the corresponding route till this node, 8) accumulated traveling distance along this route till this node, 9) a total demand of all nodes on the corresponding route, and 10) a total traveling distance of the corresponding route. The $3^{rd}$, $7^{th}$, and $9^{th}$ embedding elements are normalized by the capacity of the vehicles, and the rest are normalized by the time span of the depot $t_{max}$.

\subsubsection{Hierarchical Graphs}
\begin{figure}
    \centering
    \includegraphics[width=.6\linewidth]{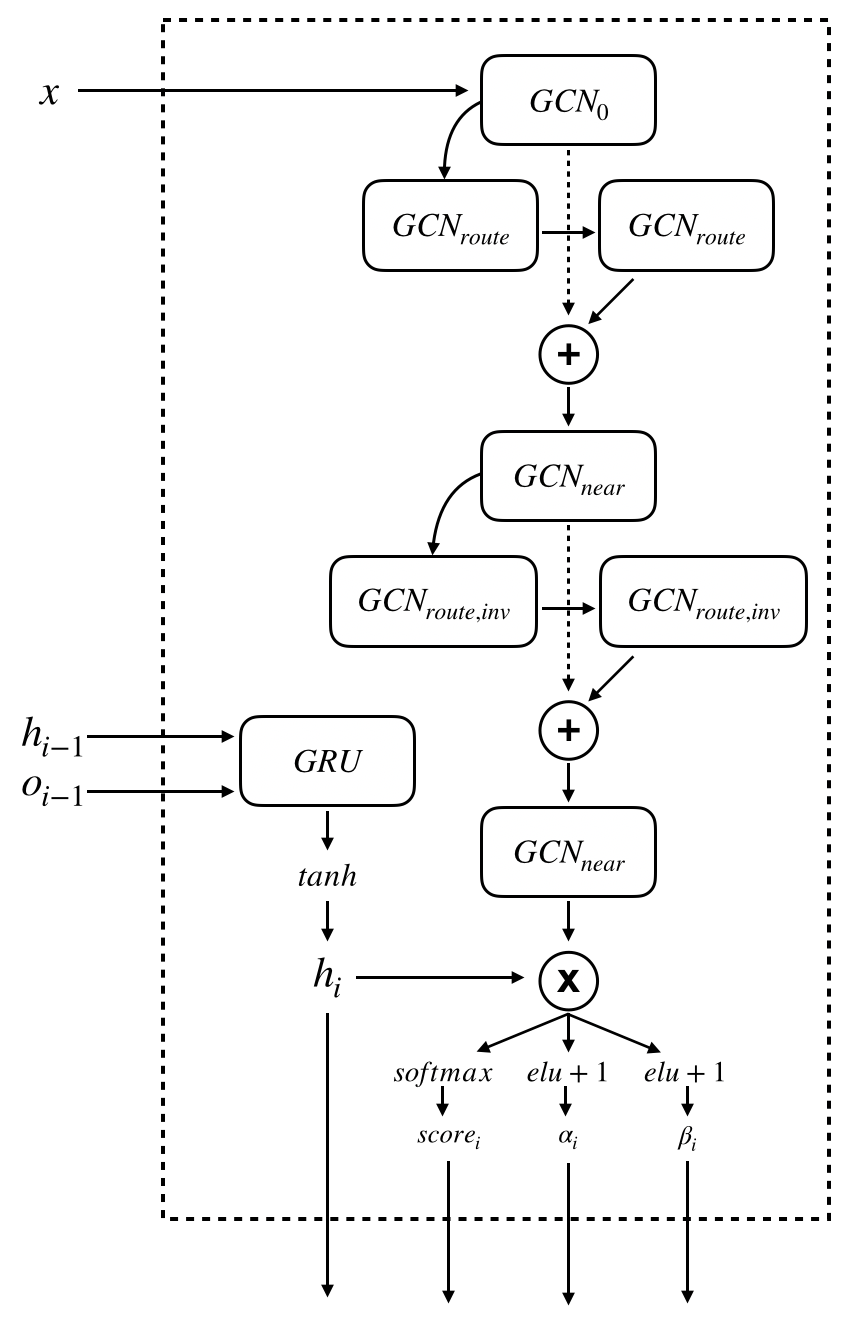}
    \caption{Illustration of the actor's network HRGCN.}\label{Fig:actor_net}
\end{figure}

The anchor node and the route coefficients are generated by a neural network namely Hierarchical Recurrent Graph Convolutional Network (HRGCN). The key ideas underlying HRGCN are the spatial propagation and the temporal propagation, as explained below. The architecture of HRGCN is illustrated as Figure \ref{Fig:actor_net}. It starts from a graph convolutional network $GCN_0$ whose nodes are the depot and customer nodes, and the arcs are the k-nearest-neighbor (k-NN) connections for each node. The input to $GCN_0$ is the 10-dimensional node embedding for each node, and the output is the enhanced node embedding with a higher dimensionality $N_A$ by integrating the information on topological contiguous nodes.  The output of $GCN_0$ is fed to two identical graph convolutional networks $GCN_{route}$ in a row, whose nodes are with the $N_A$ dimensional embedding and arcs are the segments of the routes in the current solution. This type of hierarchy $GCN_0\rightarrow GCN_{route}\rightarrow GCN_{route}$ helps spatial exploration and increases the \textit{receptive field} that is extended to longer hops among the nodes, especially along the current routes. A residual connection from $GCN_0$ to the last $GCN_{route}$ is added to prevent from gradients vanishing or exploding. This type of hierarchy is replicated consecutively with $GCN_0$ being replaced by $GCN_{near}$ and $GCN_{route}$ being replaced by $GCN_{route, inv}$, where $GCN_{near}$ is identical to $GCN_0$ except for the dimensionality of the input  node embedding  being $N_A$ instead of 10, and $GCN_{route, inv}$ is identical to $GCN_{route}$ except for the arcs being the segments of the current routes in a reversed direction. For example, given a route in current solution as $A\rightarrow B \rightarrow C \rightarrow D$, the arcs in $G_{route, inv}$ are $D \rightarrow C$, $C \rightarrow B$, and $B \rightarrow A$.  The purpose to introduce $G_{route,inv}$ is to provide a scheme about propagating node information from future nodes who are to be visited later, as a backward view.

The temporal propagation that integrates information from historical solutions is carried out by a Gated Recurrent Unit (GRU) as shown in Figure \ref{Fig:actor_net}. The input to this GRU cell is the embedding of the previous iteration's anchor node, coupled with the hidden state from the previous cell. At each iteration, the hidden state $h_i$ helps calculate the output (say, the anchor node and the route coefficients) as a multiplicative adaptive weight. 

The last layer of HRGCN outputs the probability of each node being the anchor node, by a softmax function. Two additional scalars $\alpha_i$ and $\beta_i$ are calculated by the $elu$ function for each node $i$. The parameterized Beta distribution $Beta(\alpha_i, \beta_i)$ is the probability distribution from which the route coefficients are sampled. HRGCN outputs the parameterized distribution on route coefficients rather than the deterministic coefficients values, because randomness introduced by the distribution helps the exploration efficiency.

Instead of using one anchor at each step, we have discovered from experiments that DPR with multiple anchors at each iteration usually yields better performance. Sampling multiple anchor points from this softmax function can search several areas simultaneously, thus prevents the searching process from falling into a local optima for medium-to-large scaled problems.

\begin{figure}[!hbt]
    \centering
    \includegraphics[width=.6\linewidth]{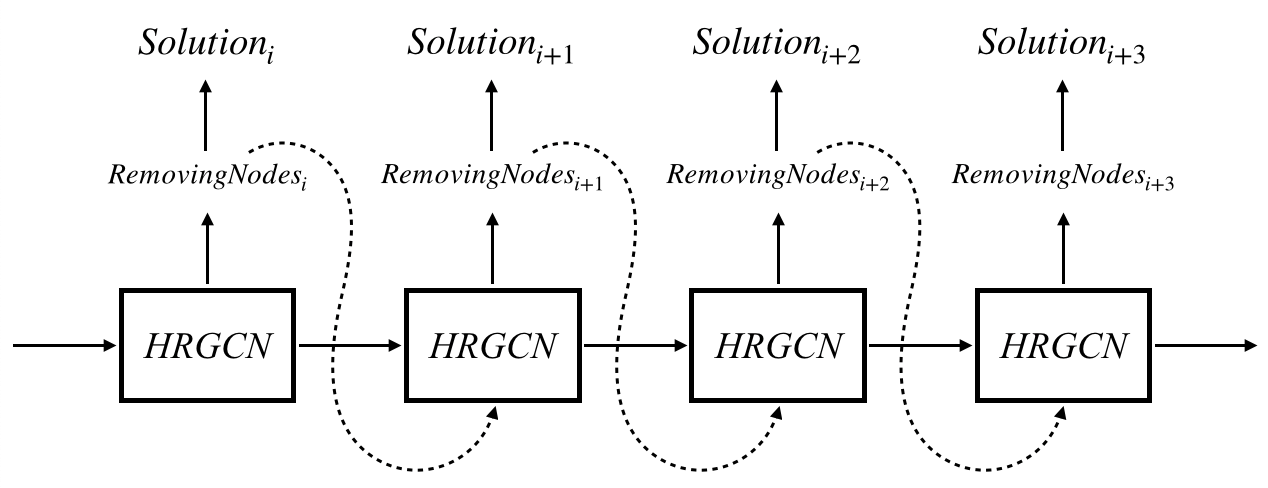}
    \caption{The recurrent perspective of HRGCN carried out via the GRU cell.}\label{Fig:hrgcn}
\end{figure}

\subsubsection{PPO for Training}

This model is trained by the Proximal Policy Optimization (PPO) framework, where the HRGCN is the actor network, and the loss function is defined as:
\begin{eqnarray}
L_A(\theta) &=& \hat{\mathbb{E}}_t[\text{min}(r_t(\theta)\delta_{\text{TD}}^{(t)}\text{, clip}(r_t(\theta), 1-\epsilon, 1+\epsilon)\delta_{\text{TD}}^{(t)})]\notag\\
L_C(\theta) &=& MSE(v, R) \notag\\
L_E(\theta) &=& Entropy(d_c) + Entropy(d_b) \notag\\
Loss(\theta) &=& -L_A(\theta)+0.5*L_C(\theta)-0.01*L_E(\theta)
\end{eqnarray}

Here $v$ represents the output of the critic network; $R$ is the discounted cumulative reward; $d_c$ is a vector of probabilities for each node being selected as the anchor node, and this vector is the output of softmax function at the last layer of HRGCN; $d_b$ denotes the beta distribution from which the route coefficient is sampled. The reward function is as simple as the delta of the cost function $r(c_{t-1}, c_{t}) = c_{t-1} - c_{t}$, where $c_{t}$ is the cost at the $t$-th iteration. 

The critic network is a multi-layer fully-connected neural network containing two hidden layers with a size of $N_C$ for both layers. The input of the critic network is the hidden state $h_i$ of the actor network, and the output is a scalar variable representing the value of the current state.

\section{Experiments}

\subsection{Problem Generation}

In this section we manifest the experiments' results of solving the CVRPTW by the proposed approach as an example. Many VRPs involve scheduling visits to customer nodes who are only available to be served during specific time windows, therefore it is a more meaningful scenario in practice. Meanwhile the tight constraints on time-windows also pose a challenge to the algorithms. 

We train the model using synthetic data sets produced by a data generator, and test it on two data sets: the synthetic data sets, as well as the Solomon benchmarks and Gehring \& Homberger benchmarks which are the \textit{de facto} benchmark data sets for VRP. The synthetic data sets are generated in a way that the positions of the customer nodes are randomly located on a map or clustered into several groups, where the number of clusters is randomly chosen from $[5, max(6, N_{nodes}/5))$. The demands are sampled from the normal distribution $\mathcal{N}(20, 121)$. If the demand is lower than 1 or higher than 45, then it will be resampled from a uniform distribution $U[5, 36]$. The service time periods for customers are identical for each instance of problem, which is randomly chosen between 10 and 100. Each customer will have a probability $P_{start}$ that the time window is not limited. For the other customers with limited time windows, the starting time $T_{Start}$ is picked from a uniform distribution $U[0, t_{max}]$, where $t_{max}$ is chosen between 600 and 10000. The end time is then calculated by $T_{End} = T_{Start}+T_{Gap}$ where $T_{Gap}$ varies from 10 to 1000.

\subsection{Setup}

We have tested the model with different settings by 1) with or without the GRU cells in the actor network and 2) by using one anchor or multiple anchors. The results verify our proposal that recurrent GRU cells that propagate information in a temporal horizon, together with multiple anchors that search more spatial areas, can notably boost the efficiency of the heuristic. Furthermore, networks trained for large-scale problems are also tested regarding their solution qualities and computational speed.

For the aforesaid purpose, 4 DPR-based reinforcement learning models are trained as DPR$_{RL-nG}$, DPR$_{RL-nGL}$, DPR$_{RL-G}$, and DPR$_{RL-n}$, where $n$ represents using multiple anchors rather than one anchor, $G$ represents using the GRU cells, and $L$ represents the model is trained with large-scale problems. More specifically, DPR$_{RL-nG}$, DPR$_{RL-G}$, DPR$_{RL-n}$ are trained with mixed-scale problems containing 25 to 200 customers, and DPR$_{RL-nGL}$ is trained with problems containing 400 to 800 customers.

\subsection{Training Details}

The initial solution is generated by: 1) removing all the nodes from the solution; 2) then by inserting each node into the solution using the least cost heuristic. The models are trained with Adam optimizer. The learning rate equals $1e-5$. In our experiments, $N_A$ = 128, $N_C$ = 256, $\gamma$ = 0.99, $K_{epoch}$ = 2, and $\epsilon_{clip}$ = 0.2. The initial and the final temperatures for simulated annealing are 100.0 and 1.0, respectively.

The degree of destruction $n_{ruin}$ is set to $\sqrt{N_{nodes}}$ in case of one anchor, or 20\% larger than $\sqrt{N_{nodes}}$ in case of multiple anchors. The number of anchors is 1 in  DPR$_{RL-G}$, 2 in DPR$_{RL-nG}$ and DPR$_{RL-n}$, and 3 in DPR$_{RL-nGL}$.

It takes 8 hours on one single Nvidia Geforce 2080Ti GPU for training. To better representing each nodes in a large-scale problem, $N_A$ and $N_C$ in DPR$_{RL-nGL}$ are doubled, and the training process takes 16 hours.

\subsection{Experiments Results}
The inference time of our model is shown as table \ref{tab:experiment_time} running on one single  Nvidia Geforce 2080Ti. For problems with no more than 200 nodes, the solving time is less than 1 second. For large-scale problem with 400 - 800 nodes, the solving time is still within a reasonable range. The speed will be faster if running in parallel on multiple GPUs.
\begin{table}[!htb]
    \centering
    \begin{tabular}{llll}
    \toprule
               & DPR & DPR & DPR  \\
    Problem    & $_{RL-nG}$ & $_{RL-G}$ & $_{RL-n}$  \\
    \midrule
    CVRPTW25  & 0.1070 & 0.1041 & 0.1026 \\
    CVRPTW50  & 0.1724 & 0.1785 & 0.1829 \\
    CVRPTW100 & 0.3690 & 0.3657 & 0.3704 \\
    CVRPTW200 & 0.9398 & 0.9316 & 0.9286 \\
    \midrule
    & DPR$_{RL-nGL}$ & \\
    \midrule
    CVRPTW400 & 6.474 & & \\
    CVRPTW600 & 13.58 & & \\
    CVRPTW800 & 20.44 & & \\
    \bottomrule
    \end{tabular}
    \caption{Inference time (time to converge to the final solution by a well trained network) for CVRPTW with different scales, on one Nvidia Geforce 2080Ti and in seconds.}\label{tab:experiment_time}
\end{table}

\begin{table*}[!ht]
    \centering
    \bigskip
    \begin{tabular}{@{}*{9}{l}@{}}
        \multicolumn{9}{@{}l}{\em(a) Experimental results of CVRPTW problems from Solomon benchmarks and Gehring \& Homberger benchmarks.}\\
        \toprule
        \multicolumn{2}{c}{ } & \multicolumn{4}{c}{Heuristic Methods} & \multicolumn{3}{c}{DL Methods} \\
        \cmidrule(r){3-6}
        \cmidrule(r){7-9}
        Problem & GT & RAND & ALNS & SISR & DPR & DPR$_{RL-nG}$ & DPR$_{RL-G}$ & DPR$_{RL-n}$ \\
        \midrule
        CVRPTW25  & 331.27 & 365.60 & 364.00 & 355.03 & 349.72 & \textbf{342.07} & 352.59 & 349.68 \\
        CVRPTW50  & 581.02 & 678.53 & 673.16 & 644.59 & 622.52 & \textbf{608.84} & 618.48 & 618.73 \\
        CVRPTW100 & 995.89 & 1196.37 & 1202.54 & 1151.82 & 1107.52 & \textbf{1067.42} & 1097.72 & 1080.49 \\
        CVRPTW200 & 2780.80 & 3698.65 & 3776.98 & 3455.53 & 3268.78 & \textbf{3107.67} & 3265.72 & 3159.38 \\
        \midrule
        & & & & & & DPR$_{RL-nG}$ & DPR$_{RL-nGL}$ & \\
        \midrule
        CVRPTW400 & 6461.14 & 9462.62 & 10134.9 & 9091.25 & 8431.59 & 8161.93 & \textbf{8131.39} & \\
        CVRPTW600 & 13078.9 & 20938.3 & 21885.0 & 19875.2 & 18221.0 & 17696.4 & \textbf{17527.8} & \\
        CVRPTW800 & 22127.9 & 36936.1 & 38327.3 & 34750.5 & 32053.7 & 30597.3 & \textbf{30498.3} & \\
        \bottomrule
    \end{tabular}
    
    \bigskip\bigskip
    \begin{tabular}{@{}*{8}{l}@{}}
        \multicolumn{8}{@{}l}{\em(b) Experimental results of CVRPTW problems on synthetic data set.}\\
        \toprule
        & \multicolumn{4}{c}{Heuristic Methods} & \multicolumn{3}{c}{DL Methods} \\
        \cmidrule(r){2-5}
        \cmidrule(r){6-8}
        Problem & RAND & ALNS & SISR & DPR & DPR$_{RL-nG}$ & DPR$_{RL-G}$ & DPR$_{RL-n}$ \\
        \midrule
        CVRPTW25  & 114.54 & 116.64 & 114.09 & 109.98 & 110.12 & \textbf{109.94} & \textbf{109.94} \\
        CVRPTW50  & 348.96 & 360.48 & 348.67 & 331.04 & \textbf{325.01} & 327.24 & 325.38 \\
        CVRPTW100 & 1170.96 & 1217.02 & 1161.89 & 1058.32 & \textbf{1042.29} & 1063.69 & 1054.02 \\
        CVRPTW200 & 4554.98 & 4632.34 & 4454.54 & 4066.86 & \textbf{3938.87} & 3945.95 & 3962.68 \\
        \midrule
        & & & & & DPR$_{RL-nGL}$ & & \\
        \midrule
        CVRPTW400 & 12261.2 & 11901.2 & 11301.2 & 9699.34 & \textbf{9358.82} & & \\
        CVRPTW600 & 29109.9 & 26528.6 & 25788.6 & 21515.2 & \textbf{20328.7} & & \\
        CVRPTW800 & 53628.2 & 48277.8 & 45645.0 & 36992.8 & \textbf{34400.8} & & \\
        \bottomrule
    \end{tabular}
    
    \bigskip
    \caption{Experimental results on Solomon benchmarks and Gehring \& Homberger benchmarks and on synthetic data sets. 150 iterations with the same solution initialization method are applied to both the heuristic methods and our models are tested. In this experiment, 4 variants of our models are presented, which are 1) DPR$_{RL-nG}$ (25-200 nodes), 2) DPR$_{RL-nGL}$ (400-800 nodes), 3) DPR$_{RL-G}$ (25-200 nodes), and 4) DPR$_{RL-n}$ (25-200 nodes)}\label{tab:experiment_result}

\end{table*}

The results are compared with: 1) the ground truth result (GT); 2) LNS with random node removal operator and least cost heuristic (RAND); 3) adaptive large heuristic search (ALNS) \cite{azi2014adaptive}; and 4) Slack Induction by String Removals (SISR) \cite{christiaens2020slack}. The initial solutions for these heuristic methods are generated in the same way as explained above. For some benchmark cases where the ground truths are not available, the best-known heuristic results are applied instead. For some 50-nodes Solomon benchmarks where both the ground truth and the heuristic results are not available, we use the SISR method running 1 million iterations as the benchmark. Table \ref{tab:experiment_result} shows that for medium-size CVRPTW problems, DPR$_{RL-nG}$ outperforms traditional heuristic search results in most cases, implying that both the recurrent structure and multiple anchors at each iteration improve the average quality of the solutions. The results also show the great potential of our model in solving large scale problems.

\begin{figure}
    \centering
    \begin{subfigure}[b]{0.475\linewidth}
        \centering
        \includegraphics[width=\linewidth]{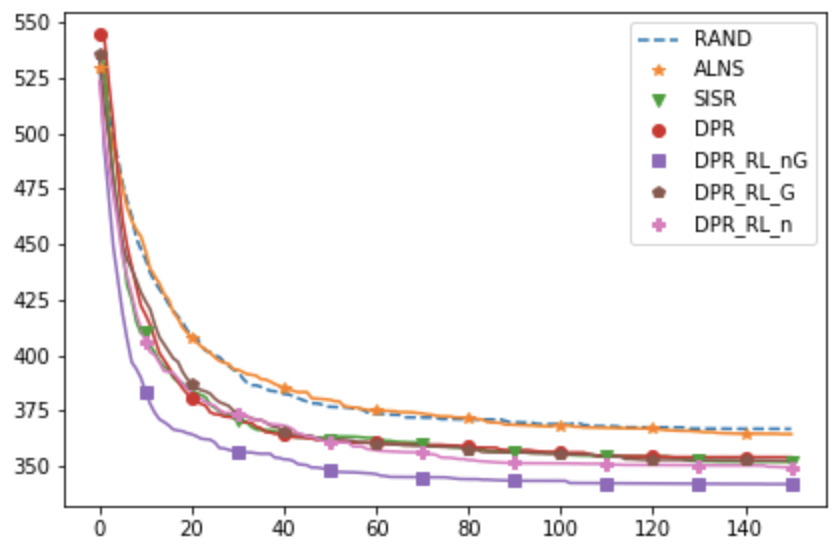}
        \caption[]%
        {{\small Scale = 25 nodes}}    
        \label{fig:descent25}
    \end{subfigure}
    \hfill
    \begin{subfigure}[b]{0.475\linewidth}  
        \centering 
        \includegraphics[width=\linewidth]{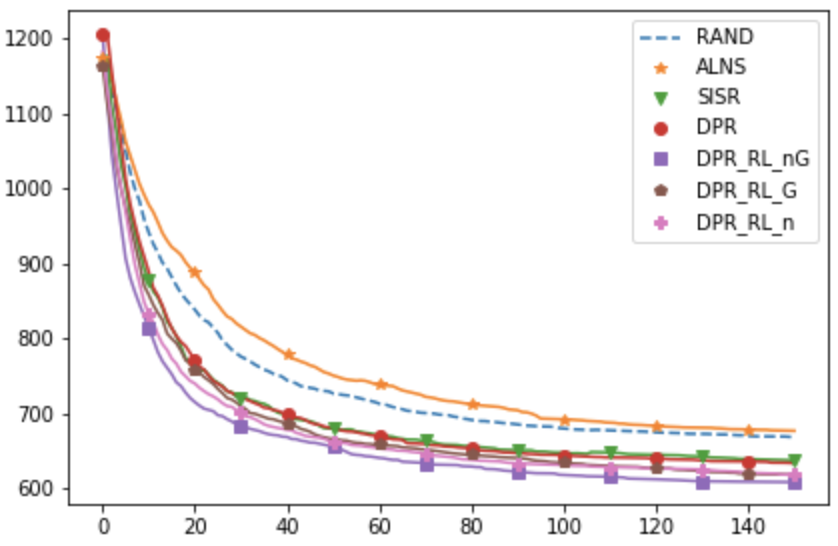}
        \caption[]%
        {{\small Scale = 50 nodes}}    
        \label{fig:descent50}
    \end{subfigure}
    \vskip\baselineskip
    \begin{subfigure}[b]{0.475\linewidth}   
        \centering 
        \includegraphics[width=\linewidth]{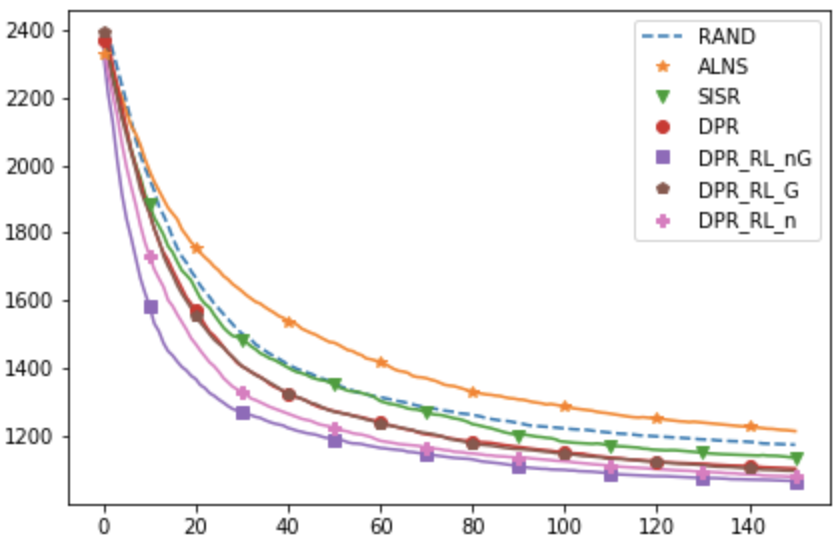}
        \caption[]%
        {{\small Scale = 100 nodes}}    
        \label{fig:descent100}
    \end{subfigure}
    \hfill
    \begin{subfigure}[b]{0.475\linewidth}   
        \centering 
        \includegraphics[width=\linewidth]{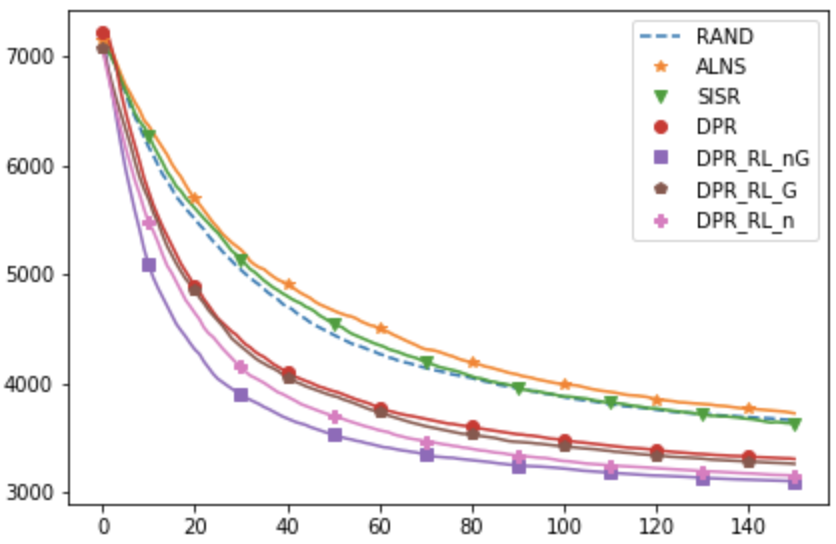}
        \caption[]%
        {{\small Scale = 200 nodes}}    
        \label{fig:descent200}
    \end{subfigure}
    \caption[]
    {\small The average costs of solutions at each iteration step using different heuristics. The results are tested on the Solomon benchmark (25-100 nodes) and the Gehring \& Homberger benchmarks (200 nodes). } 
    \label{fig:descents}
\end{figure}

Fig. \ref{fig:descents} illustrated the convergence of the averaged inference results on benchmark data sets. It shows that the proposed model has a faster converging speed than the traditional heuristic algorithms.

\begin{figure}[!hbt]
    \centering
    \includegraphics[width=.7\linewidth]{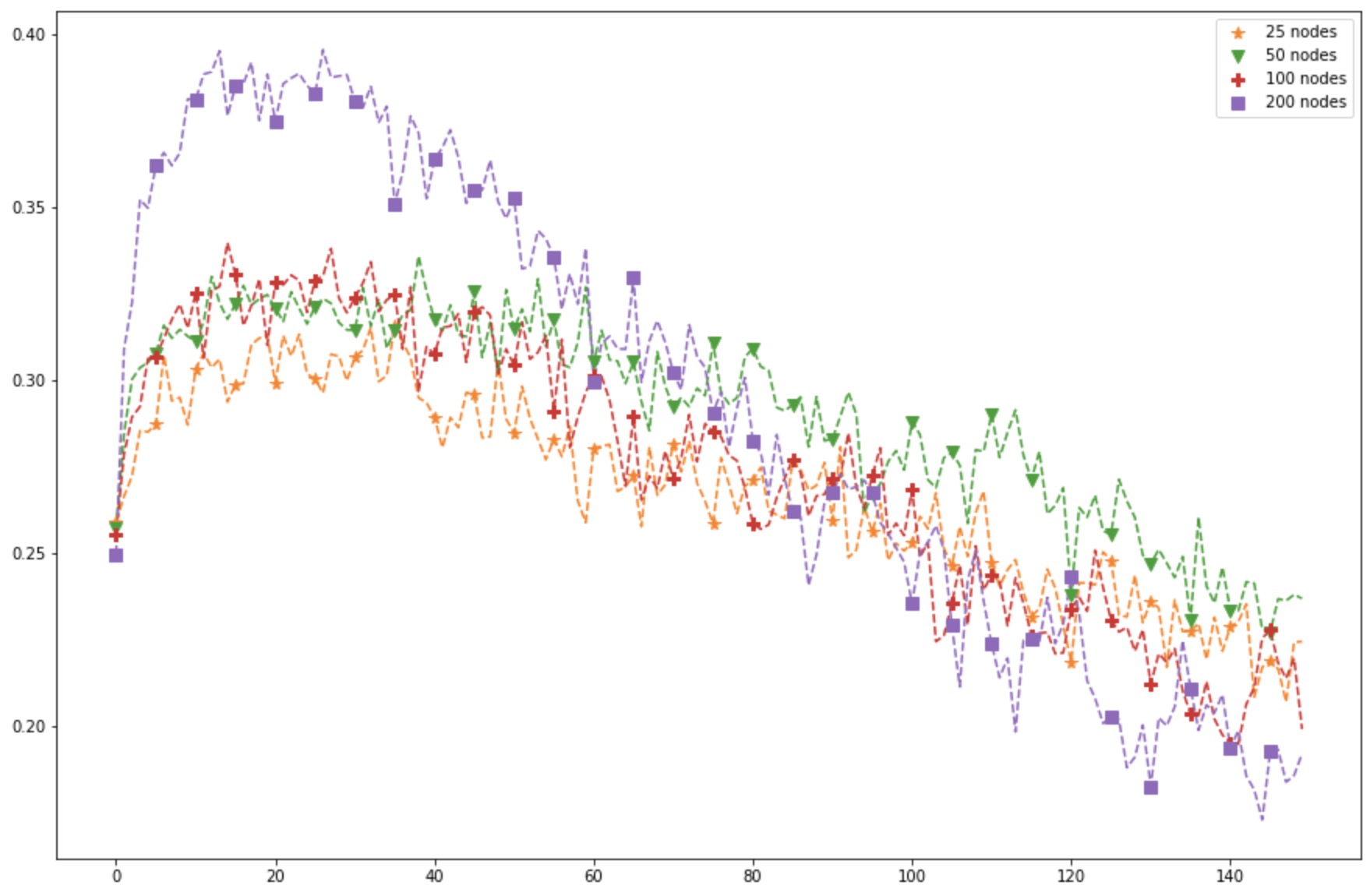}
    \caption{The averaged route coefficients per iteration step.}\label{Fig:coefficients}
\end{figure}

Fig. \ref{Fig:coefficients} shows the average route coefficients per iteration step produced by HRGCN on Solomon benchmarks (with 25, 50, 100 nodes) and on  Gehring \& Homberger benchmarks (with 200 nodes). From the plot one can conclude that the route coefficients automatically decrease along the iteration process, hence the search areas effectively zoom in from a coarse granularity to a fine one. The trend of zooming-in is more salient (say, larger initial value, deeper descending trend) for problems with larger scale, showing that this model is also adaptive to different scales.

\section{Conclusion}

In this paper, we proposed a novel approach for LNS heuristics on combinatorial optimization problems. Our model consists of an innovative neural network design namely Hierarchical Recurrent Graph Convolutional Network (HRGCN) which acts as the destroy operator of LNS heuristic in a way we introduce as Dynamic Partial Removal (DPR). DPR is inspired by Adaptive LNS (ALNS) \cite{ropke2006adaptive} about the adaptive removal selection, as well as by Slack Induction by String Removals \cite{christiaens2020slack} about the clustered-destruction insight. DPR extends the above insights in a way with more intelligence and context-awareness provided by the HRGCN, due to the fact that HRGCN is able to propagate and incorporate information from both the spatial context (the topological neighbors and adjacent nodes, as well as the contiguous nodes as a string along the routes), and the temporal context (the embedded information of the entire solutions on previous iterations). We test our model both on a set of well-known benchmarks and on synthetic data sets. The experimental results show that our model outperforms in terms of the solutions' quality and computational efficiency. Although we use CVRPTW as an example for illustration and demonstration,  the proposed model can be applied to other combinatorial optimization problems that can be structured as a graph, e.g., Network Flow problem, Generalized Assignment Problems, and other VRP variants including Multi-Depot VRP (MDVRP), VRP with Mixed Delivery \& Pickup (VRPMDP), Pick-up and delivery with time windows (PDPTW), and etc.

\bibliographystyle{unsrt}
\bibliography{./ijcai20}

\end{document}